\documentclass[lettersize,journal]{IEEEtran}
\usepackage{amsmath,amsfonts}
\usepackage{algorithmic}
\usepackage{algorithm}
\usepackage{array}
\usepackage[caption=false,font=normalsize,labelfont=sf,textfont=sf]{subfig}
\usepackage{textcomp}
\usepackage{stfloats}
\usepackage{url}
\usepackage{verbatim}
\usepackage{graphicx}
\usepackage{cite}
\usepackage{booktabs}
\usepackage{multirow}

\usepackage{pifont,xcolor}

\usepackage[colorlinks=false,      
linkcolor=black,      
citecolor=black,      
filecolor=black,      
urlcolor=blue]{hyperref}

\begin{document}

\title{
Sky-Drive: A Distributed Multi-Agent Simulation Platform for Human-AI Collaborative and Socially-Aware Future Transportation
}


\author{Zilin Huang\textsuperscript{†}, Zihao Sheng\textsuperscript{†}, Zhengyang Wan\textsuperscript{†}, Yansong Qu, Yuhao Luo, Boyue Wang, Pei Li, Yen-Jung Chen, Jiancong Chen, Keke Long, Jiayi Meng, Yue Leng, Sikai Chen\textsuperscript{*}
\thanks{The corresponding author is Sikai Chen (E-mail: sikai.chen@wisc.edu). \textsuperscript{†} These authors contributed equally to this work.}
\thanks{Zilin Huang, Zihao Sheng, Zhengyang Wan, Yuhao Luo, Boyue Wang, Pei Li, Keke Long, and Sikai Chen are with the Department of Civil and Environmental Engineering, University of Wisconsin-Madison, Madison, WI 53706, USA (E-mails: \{zilin.huang, zihao.sheng, zhengyang.wan, yuhao.luo, bwang367, pei.li, klong23, sikai.chen\}@wisc.edu). 
Yansong Qu, and Jiancong Chen are with the Lyles School of Civil and Construction Engineering, Purdue University, West Lafayette, IN 47907, USA (E-mail: \{qu120, chen5281\}@purdue.edu).
Yen-Jung Chen is with the Elmore Family School of Electrical and Computer Engineering, Purdue University, West Lafayette, IN 47907, USA (E-mail: chen4126@purdue.edu).
Jiayi Meng is with the Department of Computer Science and Engineering, The University of Texas at Arlington, Arlington, TX 76019, USA (E-mail: jiayi.meng@uta.edu).
Yue Leng is with Google, Sunnyvale, CA 94089, USA (E-mail: lengc@google.com).}}

\markboth{Journal of \LaTeX\ Class Files,~Vol.~14, No.~8, August~2021}%
{Shell \MakeLowercase{\textit{et al.}}: A Sample Article Using IEEEtran.cls for IEEE Journals}


\maketitle

\begin{abstract}
Recent advances in autonomous system simulation platforms have significantly enhanced the safe and scalable testing of driving policies. However, existing simulators do not yet fully meet the needs of future transportation research—particularly in enabling effective human-AI collaboration and modeling socially-aware driving agents. This paper introduces \textbf{Sky-Drive}, a novel distributed multi-agent simulation platform that addresses these limitations through four key innovations: (a) a distributed architecture for synchronized simulation across multiple terminals; (b) a multi-modal human-in-the-loop framework integrating diverse sensors to collect rich behavioral data; (c) a human-AI collaboration mechanism supporting continuous and adaptive knowledge exchange; and (d) a digital twin framework for constructing high-fidelity virtual replicas of real-world transportation environments. Sky-Drive supports diverse applications such as autonomous vehicle–human road users interaction modeling, human-in-the-loop training, socially-aware reinforcement learning, personalized driving development, and customized scenario generation. Future extensions will incorporate foundation models for context-aware decision support and hardware-in-the-loop testing for real-world validation. By bridging scenario generation, data collection, algorithm training, and hardware integration,  Sky-Drive has the potential to become a foundational platform for the next generation of human-centered and socially-aware autonomous transportation systems research. The demo video and code are available at: \href{https://sky-lab-uw.github.io/Sky-Drive-website/}{\textcolor{magenta}{https://sky-lab-uw.github.io/Sky-Drive-website/}}.

\end{abstract}

\begin{IEEEkeywords}
Driving Simulator, Autonomous Vehicles, Human-AI Collaboration, Multi-Agent Simulation, Digital Twin.
\end{IEEEkeywords}

\section{Introduction}
\IEEEPARstart{A}{utonomous} systems and related technologies have made significant strides in recent years, demonstrating increasing maturity in perception, decision-making, and control capabilities \cite{almaskati2024convergence,sheng2024kinematics,ma2024review,chen2023taxonomy}. As these technologies continue to advance, future transportation systems are expected to consist of diverse AI-driven intelligent agents, including autonomous vehicles (AVs), human-driven vehicles (HVs), delivery robots, flying vehicles, and smart infrastructure—each capable of perceiving the environment, making decisions, and interacting with others in real time \cite{lv2024modular}. For these agents to seamlessly integrate into traffic systems governed by humans, they must not only ensure its own safe and efficient operation, but also exhibit acute perceptiveness in recognizing human intentions and preferences (e.g., comfort-oriented speed profiles), while skillfully demonstrating social behaviors comparable to their human counterparts (e.g., yielding behaviors and adherence to implicit right-of-way conventions) \cite{liu2023towards}. This dual requirement—to understand human expectations and to coordinate with other human road users (HRUs), such as human drivers, pedestrians, and cyclists, to optimize overall system performance—constitutes what we refer to as human-AI collaboration and social awareness, respectively. Therefore, future research should transcend the isolated validation of AV safety performance toward a more comprehensive investigation of human-AI collaboration and social awareness in mixed traffic environments, where multiple types of HRUs with varying levels of autonomy interact continuously.

Validating autonomous driving technologies in real-world settings presents considerable challenges due to safety risks, limited controllability, and the scale of testing required to demonstrate reliability  \cite{kalra2016driving,feng2023dense,huang2024human}. To mitigate these barriers, the autonomous driving community has developed a variety of simulation platforms, including CARLA \cite{dosovitskiy2017carla}, AirSim \cite{shah2018airsim}, SUMO \cite{lopez2018microscopic}, Vissim \cite{ptv_vissim}, Highway-Env \cite{highway-env}, MetaDrive \cite{li2022metadrive}, SMARTS \cite{zhou2021smarts}, CarSim \cite{carsim2025} and IPG CarMaker \cite{carmaker2025}. These platforms have significantly accelerated development by providing controlled testing environments. However, they face key limitations in addressing the evolving needs of future transportation research. First, while existing simulation platforms can emulate multiple agents on a single machine using rule-based or pre-trained learning-based methods \cite{lopez2018microscopic,ptv_vissim,highway-env}, they generally do not support real-time participation of HRUs across multiple terminals. This limitation hinders the collection of authentic human behavior and HRUs' interaction data. Such data is particularly valuable for studying rare but safety-critical scenarios—for example, interactions between AVs, HVs, and pedestrians—which pose significant safety and ethical risks when collected in the real world. A distributed simulation platform that enables participants to assume diverse roles across multiple terminals is urgently needed to safely collect such interaction data and to evaluate interaction algorithms in controlled environments. 


Second, existing simulation platforms offer limited support for human-AI collaboration. While they can collect human inputs, these are often treated as low-level control signals rather than high-level feedback for improving autonomous driving algorithms \cite{gulino2023waymax,li2023scenarionet,vinitsky2022nocturne}. In contrast, effective human-AI collaboration refers to a bidirectional process in which humans provide feedback not only as commands, but also as indications of preferences, situational understanding, and normative behaviors; AI systems, in turn, assist human drivers by offering real-time guidance, performance feedback, and personalized training. This bidirectional exchange enables AI systems to continuously adapt to human needs and expectations while simultaneously enhancing human driving performance through intelligent support. Additionally, the emergence of foundation models—such as large language models (LLMs) \cite{openai2023gpt4,dubey2024llama} and vision-language models (VLMs) \cite{Qwen2VL}—trained on large-scale, multimodal datasets and equipped with broad world knowledge—offers new opportunities for capturing and utilizing human knowledge \cite{liao2024gpt}. Yet, in most simulators \cite{yang2024drivearena,zhang2024chatscene,wei2024editable}, such models are used primarily for scenario generation rather than as active components in human-AI collaborative learning. 

\begin{figure*}[!t]
\centering
\includegraphics[width=0.95\textwidth]{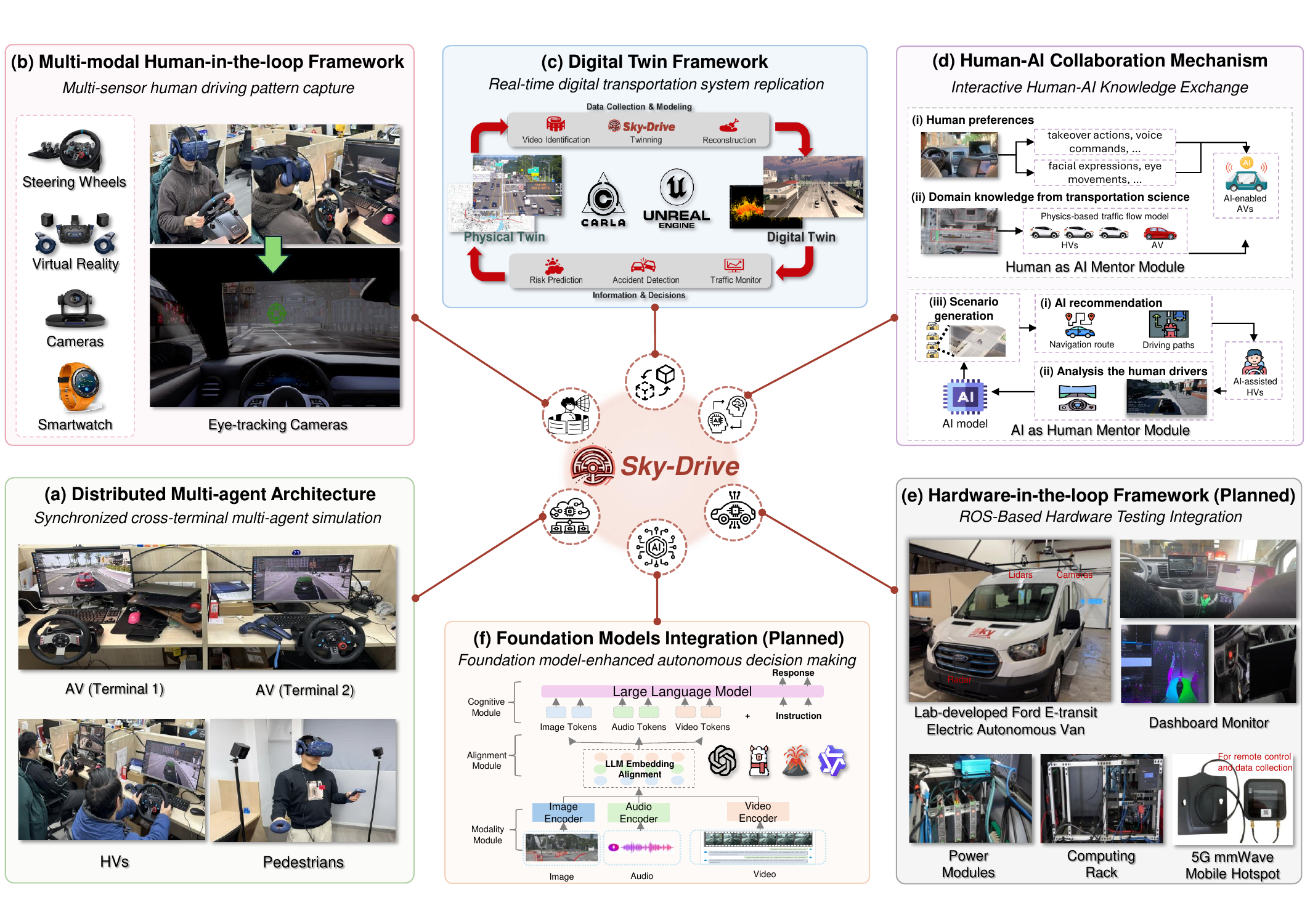}
\caption{Overview of Sky-Drive's key components and functionalities. (a) a distributed multi-agent architecture enabling synchronized simulation across multiple terminals; (b) a multi-modal human-in-the-loop framework capturing comprehensive behavioral data through integrated sensor systems; (c) a digital twin  framework that creates high-fidelity virtual replicas of transportation systems through multi-source data integration; (d) a human-AI collaboration mechanism facilitating knowledge exchange between humans and AI systems; (e) the planned integration of foundation models to enhance decision-making, enabling more adaptive and context-aware human-AI collaboration; (f) a hardware-in-the-loop framework, planned for future integration, enabling remote control and data collection and ensuring that algorithms are evaluated in real-world environments.}
\label{fig1}
\end{figure*}


Third, although some simulation platforms have integrated reinforcement learning (RL) capabilities to improve autonomous driving policies \cite{li2022metadrive,zhou2021smarts,berta2024development}, they remain largely focused on optimizing vehicle-level metrics such as safety, efficiency, and route completion. However, advancing real-world deployment requires moving beyond individual vehicle performance to incorporate social awareness into the decision-making process. Social awareness refers to an autonomous system's ability to coordinate with other traffic participants to enhance overall system performance \cite{wang2022social}. This includes promoting traffic flow stability, mitigating congestion through cooperative maneuvers, enhancing the collective comfort of all road users, and enabling harmonious interactions between AVs and humans in mixed traffic settings. In this context, transportation science offers a valuable foundation. Decades of research have produced validated traffic flow theories and human behavioral models—such as the intelligent driver model (IDM) \cite{treiber2000congested} and the minimizing overall braking induced by lane changes (MOBIL) \cite{kesting2007general} model—that can inform the design of socially aware autonomous systems capable of optimizing not just individual performance but the efficiency and safety of the entire traffic ecosystem.

To address these challenges, this paper proposes \textbf{Sky-Drive}, an open-source simulation platform designed to advance research in socially aware autonomous driving and human-AI collaboration. Sky-Drive integrates scenario generation, data collection, algorithm training, and hardware integration into a unified platform, supporting distributed multi-agent operation and multi-modal human-in-the-loop interaction. As illustrated in Fig. \ref{fig1}, Sky-Drive introduces four key innovations:
\begin{itemize}
    \item Sky-Drive introduces a distributed multi-agent architecture that enables synchronized simulation across multiple devices through remote procedure call (RPC). This design allows independent control of agents on separate terminals while maintaining shared environmental states, better reflecting future mixed traffic.
    \item Sky-Drive provides a multi-modal human-in-the-loop framework that integrates diverse sensors, including steering wheels, virtual reality (VR) systems, cameras, and smartwatches, to capture rich human behavioral data. A synchronized data processing pipeline correlates these multi-modal streams, enabling detailed analysis of human driving patterns and responses to complex scenarios.
    \item Sky-Drive implements an innovative human-AI collaboration mechanism comprising a human as AI mentor (HAIM) module that incorporates human feedback and domain knowledge to guide AI learning, and an AI as human mentor (AIHM) module that provides real-time guidance and personalized training to human drivers.
    \item  To bridge the gap between simulation and reality, Sky-Drive includes a digital twin (DT) framework that builds high-fidelity virtual replicas of transportation systems by integrating data collected from lab-developed AVs, roadside sensors, traffic cameras, and historical records.
\end{itemize}

To further enhance Sky-Drive's capabilities, two major functionalities are planned:
\begin{itemize}
    \item Sky-Drive will integrate foundation models, i.e., LLM and VLM, at both the system and agent levels. At the system level, foundation models will provide global observation and feedback to optimize simulation dynamics. At the agent level, they will enhance situational understanding and enable safer, more socially aware, and personalized decision-making.
    \item Sky-Drive will incorporate a hardware-in-the-loop (HIL) framework via robot operating system (ROS), enabling direct validation of autonomous driving algorithms on physical vehicles and safe evaluation of algorithms without exposing users to real-world risks.
\end{itemize}

The remainder of this paper is organized as follows: Section \ref{Related Works} reviews related work in driving simulators. Section \ref{Sky-Drive Framework} introduces Sky-Drive's workflow. Section \ref{Sky-Drive Features} details Sky-Drive's features and technical implementation. Section \ref{Application} demonstrates application examples. Section \ref{future} discusses planned future enhancements. Finally, Section \ref{Conclusions} concludes the paper and outlines future research directions.

\section{Related Work}
\label{Related Works}

\subsection{Driving Simulators}
Driving simulation platforms have evolved significantly to address the growing needs of AVs research. According to Li et al. \cite{li2024choose}, these simulators can be categorized based on their primary functions and capabilities.

Comprehensive simulators provide end-to-end virtual environments with complete road networks, diverse traffic agents, pedestrians, and detailed sensor models. CARLA \cite{dosovitskiy2017carla} and LGSVL \cite{rong2020lgsvl} represent prominent open-source examples in this category, offering rich environments for testing autonomous driving systems. Commercial solutions such as Nvidia Drive Sim \cite{nvidia2024drive} and rFpro \cite{rfpro2023simulation}, alongside academic developments including DeepDrive \cite{team2019deepdrive} and GarchingSim \cite{zhou2023garchingsim}, provide similar comprehensive capabilities. Another important category is traffic flow simulators, which focus on modeling network-level vehicle movements, traffic congestion, and large-scale traffic scenarios. Notable examples include SUMO \cite{lopez2018microscopic}, Vissim \cite{ptv_vissim}, Flow \cite{wu2021flow}, and CityFlow \cite{zhang2019cityflow}. Recent developments combine SUMO's traffic modeling with 3D simulators such as CARLA to merge scalability with realism.

Sensory data simulators, such as AirSim \cite{shah2018airsim} and Sim4CV \cite{muller2018sim4cv}, are designed to generate high-fidelity sensor outputs for perception systems. These functionalities are increasingly being integrated into comprehensive simulators while maintaining their critical role in AV perception testing. Driving policy simulators provide configurable environments for evaluating decision-making algorithms. Examples include Highway-Env \cite{highway-env}, TORCS \cite{wymann2020torcs}, SUMMIT \cite{cai2020summit}, MACAD \cite{palanisamy2020multi}, SMARTS \cite{zhou2021smarts}, and MetaDrive \cite{li2022metadrive}. Additionally, recent data-driven simulators such as Waymax \cite{gulino2023waymax}, ScenarioNet \cite{li2023scenarionet}, and Nocturne \cite{vinitsky2022nocturne} leverage real-world datasets to generate socially relevant traffic scenarios. Vehicle dynamics simulators, including CarSim~\cite{carsim2025}, IPG CarMaker~\cite{carmaker2025}, and Gazebo~\cite{gazebo2025}, specialize in accurately modeling vehicle physics, such as suspension responses and tire-road interactions, which are essential for validating control algorithms under realistic conditions.

While existing platforms offer valuable simulation capabilities, certain challenges remain in supporting future transportation research. As shown in Tab.~\ref{tab1}, most simulators run only on single devices, limiting their ability to model distributed multi-agent scenarios. Additionally, current platforms provide insufficient support for socially-aware algorithms that need to understand complex interactions with diverse road users. Considering CARLA's established strengths in sensor simulation and visualization, Sky-Drive leverages CARLA as its core engine while extending it with a distributed architecture for synchronized multi-terminal simulation, immersive VR interfaces, and a DT framework. This integration creates an open-source platform specifically designed to support future transportation research.

\begin{table*}[ht]
\centering
\caption{Comparison of Representative Simulators with Sky-Drive}
\resizebox{\textwidth}{!}{%
\begin{tabular}{lcccccc}
\toprule
\textbf{}                      & \textbf{Distributed}            & \textbf{Digital Twin}      & \textbf{Hardware-}          & \textbf{Traffic Flow}     & \textbf{AI Framework}     & \textbf{Human-in-the-} \\
& \textbf{Multi-agent Simulation} & \textbf{Environment}       & \textbf{in-the-Loop}        & \textbf{Modeling}         & \textbf{Integration}      & \textbf{loop Interface} \\
\midrule
\multicolumn{7}{c}{\textbf{Closed Source}} \\ 
\midrule
Nvidia Drive Sim  \cite{nvidia2024drive}     & -                    & \checkmark            & \checkmark                    & -                     & \checkmark               & \checkmark                \\ 
rFpro \cite{rfpro2023simulation}                 & -                    & \checkmark            & \checkmark                    & -                     & -                        & \checkmark                \\ 
CarSim  \cite{carsim2025}               & -                    & -                     & \checkmark                    & -                     & -                        & \checkmark                \\ 
Matlab \cite{matlab2025}                 & -                    & \checkmark            & \checkmark                    & -                     & \checkmark               & \checkmark                \\ 
\midrule
\multicolumn{7}{c}{\textbf{Open Source}} \\ 
\midrule
DeepDrive 2.0 \cite{team2019deepdrive}         & -                    & -                     & -                              & -                     & \checkmark               & -                         \\ 
GarchingSim \cite{zhou2023garchingsim}           & \checkmark           & -                     & \checkmark                    & -                     & \checkmark               & \checkmark                \\ 
CARLA  \cite{dosovitskiy2017carla}                & -                    & \checkmark            & \checkmark                    & -                     & \checkmark               & \checkmark                \\ 
SUMO  \cite{lopez2018microscopic}                 & -                    & \checkmark            & \checkmark                    & \checkmark            & -                        & -                         \\ 
Flow  \cite{wu2021flow}                 & -                    & -                     & -                              & \checkmark            & \checkmark               & -                         \\ 
CityFlow  \cite{zhang2019cityflow}             & -                    & -                     & -                              & \checkmark            & \checkmark               & -                         \\ 
TORCS \cite{wymann2020torcs}                 & -                    & -                     & -                              & -                     & \checkmark               & -                         \\ 
SUMMIT \cite{cai2020summit}                & -                    & \checkmark            & -                              & -                     & \checkmark               & -                         \\ 
MACAD  \cite{palanisamy2020multi}                & -                    & -                     & -                              & -                     & \checkmark               & \checkmark                \\ 
MetaDrive  \cite{li2022metadrive}            & -                    & \checkmark            & -                              & -                     & \checkmark               & \checkmark                \\ 
SMARTS \cite{zhou2021smarts}                & -                    & -                     & -                              & -                     & \checkmark               & -                         \\ 
Nocturne \cite{vinitsky2022nocturne}              & -                    & \checkmark            & -                              & \checkmark            & \checkmark               & -                         \\ 
Waymax  \cite{gulino2023waymax}               & -                    & \checkmark            & -                              & \checkmark            & \checkmark               & -                         \\ 
Gazebo \cite{gazebo2025}                & -                    & \checkmark            & -                              & -                     & -                        & \checkmark                \\ 
\textbf{Sky-Drive (Ours)} & \checkmark           & \checkmark            & \checkmark                    & \checkmark            & \checkmark               & \checkmark                \\ 
\bottomrule
\end{tabular}%
}
\par\vspace{1em}
\raggedright \small Note: The ``Distributed Multi-agent Simulation" functionality in this table refers to the capability of simulators to synchronize and run multiple agents (e.g., AVs, HVs, and pedestrians) across different computers in real-time simulations. This is distinct from simply running multiple agents concurrently on a single computer, which most simulators can accomplish.
\label{tab1}
\end{table*}

\subsection{Human–AI Collaboration Environments}
Several simulation platforms have contributed to advancing human-AI collaboration in autonomous driving. For instance, NVIDIA's DRIVE Sim and Omniverse platform \cite{nvidia2025drive} support collaboration by generating physics-based synthetic data for training autonomous systems. However, their approach largely enables one-way knowledge transfer—where simulated data informs AI models—without supporting real-time, bidirectional human-AI interaction. Applied Intuition \cite{applied2025intuition} incorporates human-in-the-loop testing to allow operators to validate autonomous decisions, yet its framework is primarily tailored for offline validation rather than continuous learning. MORAI \cite{morai2025} provides digital twin environments that visualize AI decision-making for human drivers, but its interaction remains limited to basic feedback collection without mechanisms for mutual adaptation or learning.

More specialized platforms have made progress toward collaborative learning. MIT’s VISTA \cite{amini2022vista} enables domain adaptation between real and virtual environments, but focuses mainly on perception rather than interactive decision-making. The GAMMA framework \cite{luo2022gamma} introduces mixed-reality traffic incorporating human behavior, though it lacks explicit mechanisms for integrating human experience into AI learning. Wayve’s LINGO architecture \cite{Wayve2023LINGO} enhances transparency by providing natural language explanations for AI decisions, while SafeMod \cite{ma2024bidirectional} leverages LLMs for bidirectional planning, mimicking human reasoning in autonomous decision-making. Similarly, SurrealDriver \cite{jin2023surrealdriver} generates realistic driving behaviors that align with human expectations using LLMs, and DarwinAI’s GenSynth \cite{shafiee2019human} facilitates collaboration between human designers and AI in developing neural networks for driving tasks.

Despite these advances, most platforms still fall short in enabling true human-AI knowledge exchange. They often lack mechanisms for the continuous integration of human feedback, resulting in open-loop rather than closed-loop learning processes. Moreover, few platforms support comprehensive multimodal sensing from humans—such as gaze, voice, physiological signals, and control inputs—which are critical for modeling and understanding driving behaviors. Sky-Drive addresses these limitations through its HAIM and AIHM modules, its multi-modal human-in-the-loop framework, and its closed-loop learning architecture that continuously integrates human experience into AI development.

\begin{figure*}[ht]
\centering
\includegraphics[width=0.9\textwidth]{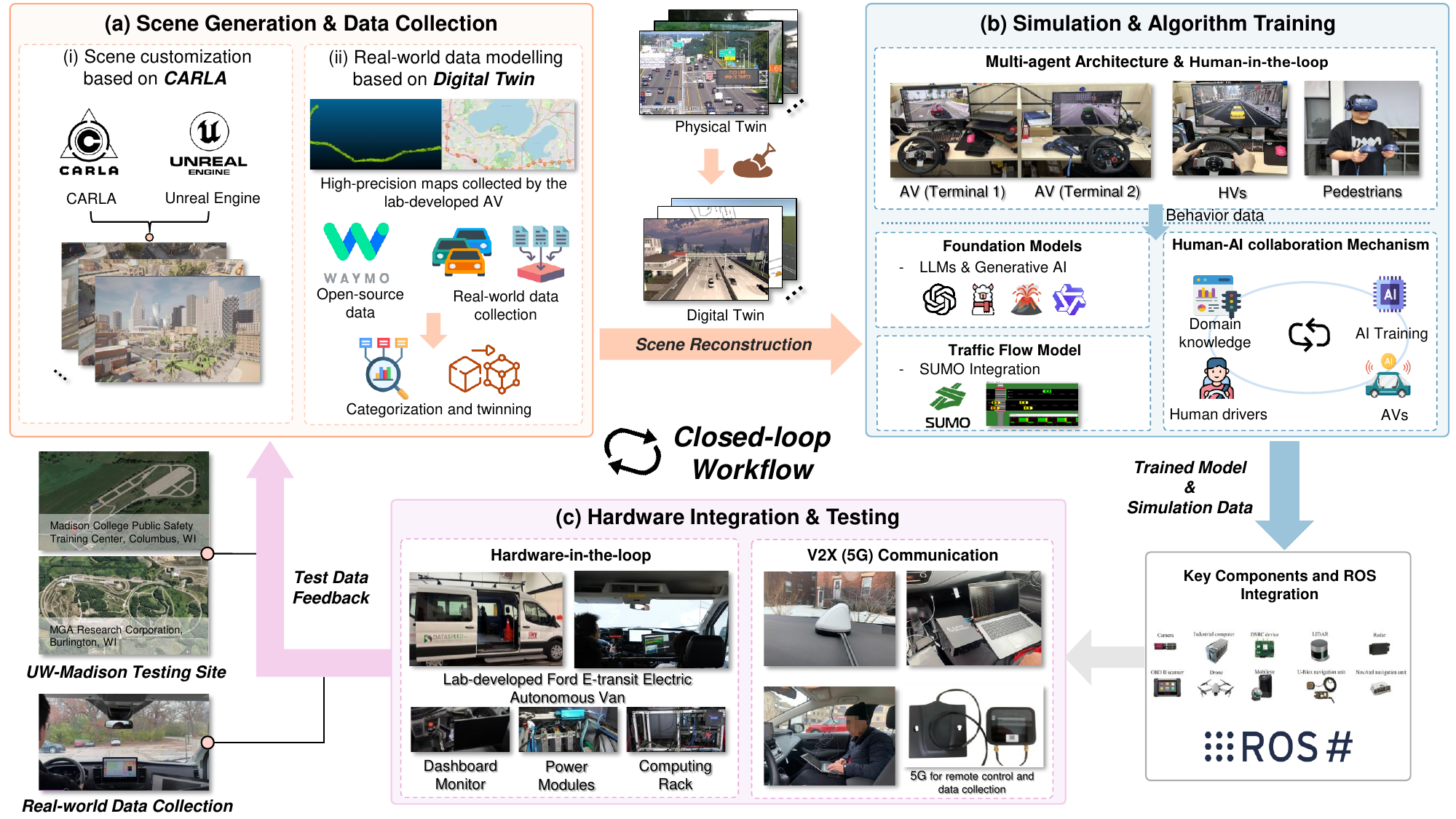}
\caption{Workflow of Sky-Drive.  (a) scenario generation \& data collection through CARLA-based synthetic environments and digital twin integration of real-world traffic data; (b) simulation \& algorithm training enabled by distributed multi-agent architecture and human-AI collaboration mechanism; (c) hardware integration \& testing utilizing ROS compatibility for direct validation of autonomous driving algorithms on physical platforms.}
\label{fig2}
\end{figure*}

\section{Sky-Drive Workflow}
\label{Sky-Drive Framework}
\subsection{Overview}

As shown in Fig. \ref{fig1}, the workflow begins with \ding{182} the \textit{DT framework}, which creates high-fidelity virtual replicas of transportation systems through multi-source data integration. These virtual environments feed into \ding{183} the \textit{distributed multi-agent architecture}, enabling synchronized simulations across multiple devices and supporting complex interactions between autonomous agents. Together, the DT framework and the multi-agent architecture form the simulation environment that serves as the testing ground for \ding{184} the \textit{multi-modal human-in-the-loop framework}, which captures comprehensive behavioral data from human participants. \ding{185} The \textit{human-AI collaboration mechanism} then processes and utilizes this data, facilitating knowledge exchange between humans and autonomous systems. \ding{186} The \textit{foundation models} integration will enhance system and agent-level capabilities, enabling observations for performance feedback and aiding individual agents in better understanding human behavior patterns. Finally, \ding{187} the \textit{HIL framework} connects with the DT framework, enabling real-world algorithm validation while feeding real-world performance data back into the simulation.

\subsection{Details}
Sky-Drive's detailed workflow, shown in Fig. \ref{fig2}, consists of three primary stages that form a continuous feedback loop:

\subsubsection{Scenario Generation \& Data Collection}
As shown in Fig. \ref{fig2} (a), this stage employs two complementary approaches to ensure comprehensive scenario coverage: (i) Sky-Drive leverages CARLA and Unreal Engine to generate customizable urban environments with detailed road networks, traffic rules, and environmental conditions, enabling controlled testing of specific driving scenarios. (ii) The DT framework imports real-world data through multi-source integration, including high-definition maps collected by lab-developed AVs, open-source data, and real-world traffic data collection. This data undergoes categorization and twinning processes to create digital replicas of physical environments.

\subsubsection{Simulation \& Algorithm Training}
As shown in Fig. \ref{fig2} (b), this stage processes the generated scenarios through an integrated learning pipeline with four interconnected components: (i) The distributed multi-agent architecture enables the concurrent operation of multiple agents across different terminals, facilitating complex traffic interactions in a shared environment while maintaining synchronization. (ii) The human-in-the-loop component integrates multiple human participants, capturing human behavior through immersive interfaces and allowing for real-time feedback collection. (iii) Sky-Drive will integrate LLMs to enhance simulation capabilities, facilitating natural communication between human participants and autonomous systems for more intuitive interaction and knowledge transfer. (iv) The human-AI collaboration mechanism integrates human feedback and domain knowledge into AI training, creating a continuous learning loop where humans inform AI systems and AI provides feedback to humans.

\subsubsection{Hardware Integration \& Testing}
As illustrated in Fig. \ref{fig2} (c), this stage bridges simulation and physical deployment through two key components: (i) While the full HIL framework is planned for future development, the current architecture already supports connections to external hardware through standardized ROS. The lab-developed Ford E-transit electric van serves as the primary testbed, equipped with dashboard monitors, power modules, drive-by-wire systems, and a computing rack for algorithm deployment. Testing is primarily conducted at the Madison College Public Safety Training Center in Columbus, WI, and MGA Research Corporation in Burlington, WI. (ii) Sky-Drive also integrates a 5G mmWave mobile hotspot to support low-latency teleoperation, enabling remote human operators to monitor and, when necessary, take control of physical vehicles in real time. 

\subsection{Case Demonstration}
To demonstrate Sky-Drive’s workflow, we consider the case of personalized autonomous driving. In this case, Sky-Drive develops an autonomous system that tailors its behavior to the driver’s unique preferences, learning from their driving styles and comfort levels. 

The workflow begins with the DT framework, which creates high-fidelity virtual replicas of real-world traffic environments using data such as high-definition maps. These environments are then input into the distributed multi-agent architecture, enabling simulations of complex interactions between AVs and humans. During the simulation and training stage, real-time feedback from the driver, such as “It’s too fast” or “The acceleration is too harsh,” is processed by LLMs to infer preferences regarding acceleration and driving style. This feedback is integrated into the human-AI collaboration mechanism, forming a continuous learning loop where the system adapts its driving strategies and provides more personalized guidance. The HIL framework connects the system to physical platforms, validating the personalized driving algorithm in real-world scenarios. This closed-loop workflow enables the development and validation of personalized autonomous systems, from concept testing to real-world deployment, ensuring safety and reliability.

\section{Sky-Drive Features}
\label{Sky-Drive Features}

\begin{figure*}[ht]
\centering
\includegraphics[width=0.9\textwidth]{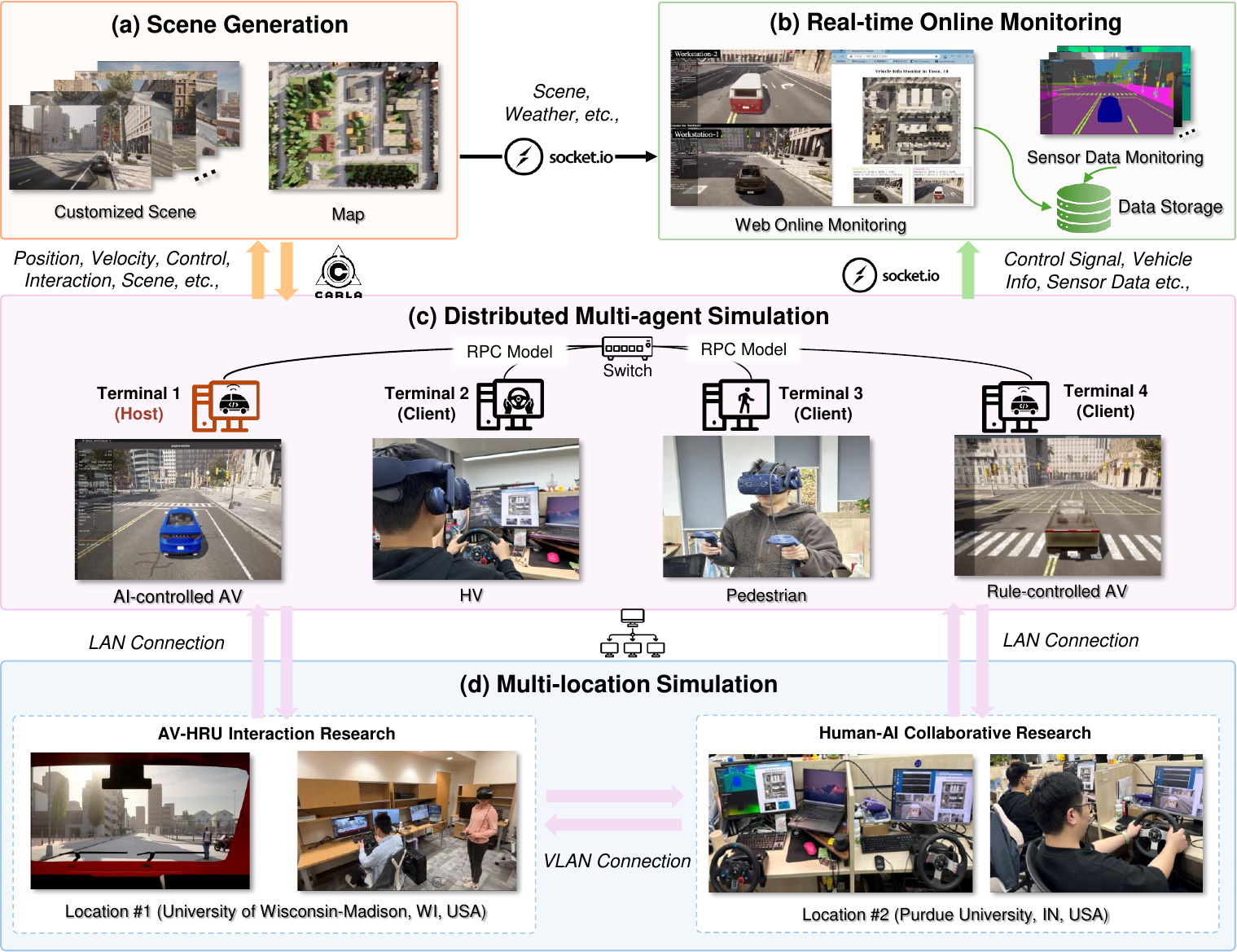}
\caption{Illustration of Sky-Drive's distributed multi-agent architecture. Sky-Drive enables synchronized simulation across multiple terminals while maintaining precise real-time interactions between AVs, HVs, and pedestrians through RPC and Socket.IO-based communication platform, supporting comprehensive data collection and real-time analysis of multi-agent behaviors.}
\label{fig3}
\end{figure*}

\subsection{Distributed Multi-agent Architecture}
Future transportation systems will consist of multiple intelligent agents, such as AVs, HVs, and pedestrians, each operating independently, requiring simulation systems that can model and control these agents separately while ensuring seamless interaction between them. Existing platforms, such as Nocturne~\cite{vinitsky2022nocturne}, MetaDrive~\cite{li2022metadrive}, and Waymax~\cite{gulino2023waymax}, primarily focus on simplified multi-agent interactions on a single machine, limiting their ability to model such complexity. To address this, Sky-Drive introduces a novel distributed multi-agent architecture that enables the synchronized simulation of multiple independently operating agents across different computing devices. 

\subsubsection{System Architecture}
At the core of Sky-Drive lies RPC built upon CARLA, using the \texttt{rpclib} library. This extension enhances CARLA's vehicle control system, enabling crucial improvements for distributed multi-agent simulation. As shown in Fig. \ref{fig3} (c), Terminal 1 functions as the host (server) that maintains the global simulation environment, while Terminals 2-4 act as clients controlling different agent types. Each terminal independently controls its corresponding agent through various input devices, while ensuring seamless interaction with other agents in the shared environment. The host terminal is responsible for scene customization and map generation, which is then distributed to the client terminals. The scene generation component (Fig. \ref{fig3} (a)) creates detailed virtual environments with customizable traffic conditions, weather patterns, and road infrastructure, supporting multiple agent types, including AI-controlled AVs, HVs, pedestrians controlled via VR, and rule-based AVs following predefined behaviors (Fig. \ref{fig1} (a)).

\subsubsection{Communication Infrastructure}
The communication infrastructure employs a dual-port TCP system on each terminal, enabling robust bidirectional data exchange between the host and clients. Sky-Drive’s hybrid networking approach ensures optimal performance. For time-critical operations, Sky-Drive utilize a dedicated local area network (LAN) with high-performance switches and Ethernet connections, achieving low latency of 0.3 milliseconds for smooth real-time interactions among agents. For scenarios requiring broader network coverage, such as geographically distributed research across Purdue University and University of Wisconsin-Madison (Fig. \ref{fig3} (d)), virtual LAN (VLAN) configurations extend the platform’s reach while maintaining communication efficiency.

\subsubsection{Real-time Monitoring Webstie}
A key component of Sky-Drive's distributed architecture is its real-time monitoring and data management website. Complementing the core communication infrastructure, Sky-Drive has developed a Socket.IO-based communication platform that tracks agent data, including position coordinates, velocity metrics, live video feeds, and sensor readings. As shown in Fig. \ref{fig3} (b), the platform features a web-based system that provides real-time visualization of agent activities. It streams data to a centralized system where agent interactions are monitored and analyzed in real-time. All simulation data, including agent states, environmental conditions, and interaction events, are logged in a centralized database, enabling comprehensive post-simulation analysis and scenario reproduction.

\subsection{Multi-modal Human-in-the-loop Framework}
To capture human preferences and cognitive states for adaptive AI behavior, Sky-Drive develops a multi-modal human-in-the-loop framework, illustrated in Fig. \ref{fig1} (b), that collects and synchronizes gaze patterns, voice commands, facial expressions, physiological signals, and control actions across multiple modalities. 

\subsubsection{Eye Tracking}
Sky-Drive provides an immersive experience through a custom-developed VR interface built on top of the Unreal Engine. Participants engage in the simulation using an HTC Vive Pro Eye headset, which supports full six degrees of freedom (6-DoF) head tracking via SteamVR and integrated eye tracking via the SRanipal SDK. The system captures high-frequency (up to 120 Hz) behavioral signals, including 3D gaze vectors, pupil positions and diameters, eye openness, and fixation points. These signals are critical for analyzing driver attention distribution, situational awareness, and cognitive workload during complex driving tasks.

\subsubsection{Voice Interaction}
Sky-Drive supports voice commands as an explicit behavioral input modality. Spoken language is transcribed via Whisper, an OpenAI automatic speech recognition (ASR) model \cite{radford2023robust}, and then interpreted by LLMs. The system extracts driver intent and sentiment from structured commands (“slow down at the next intersection”) and informal feedback (“too fast”), mapping them into semantic driving directives or policy preferences to guide AI behavior.

\subsubsection{Facial Expression Recognition}
A high-resolution in-cabin camera captures facial micro-expressions in real time. Sky-Drive employs expression classification models trained on affective datasets to recognize expressions such as stress, confusion, or satisfaction. These cues serve as implicit indicators of driver state and comfort, enabling real-time adaptation of AI behavior and intervention when necessary.

\subsubsection{Physiological Signal Monitoring}
Physiological states such as stress and alertness are inferred through biometric signals collected by wearable devices. Sky-Drive integrates the Garmin vívoactive 5 smartwatch to continuously monitor heart rate and heart rate variability (HRV). These physiological signals are synchronized with other behavioral data streams, providing additional channels to model driver arousal, cognitive workload, and fatigue.

\subsubsection{Steer Wheel}
The ego vehicle is equipped with a Logitech G920 racing wheel and pedal system, with force feedback enabled through the open-source Logitech Wheel Plugin. Steering, throttle, braking, and signaling inputs are logged in parallel with gaze and head pose data. This setup supports realistic driving control and is fully compatible with CARLA’s ScenarioRunner for scenario-based experiments.

\subsection{Human-AI Collaboration Mechanism}
Sky-Drive implements an adaptive human-AI collaboration mechanism that enables continuous, bidirectional knowledge exchange between human users and AI-enabled autonomous systems. As shown in Fig.\ref{fig1} (d), this mechanism is built on two complementary modules: HAIM and AIHM.

\subsubsection{Human as AI Mentor}
In the HAIM, humans serve as real-time mentors to AVs, guiding AI learning through two key sources of human knowledge: (i) Individual behavioral knowledge, encompassing both explicit behaviors (e.g., takeovers, voice commands, touchscreen interactions) and implicit signals (e.g., facial expressions, eye movements, physiological responses), captured via Sky-Drive’s multi-modal interface \cite{huang2024human,huang2024vlm}; (ii) Domain knowledge from transportation science, including established models such as IDM and MOBIL that encode long-standing rules of human driving behavior \cite{huang2024trustworthy}.

The HAIM adopts an RL paradigm enhanced by human preference modeling and physics-informed priors to incorporate this dual-source knowledge. Rather than relying on handcrafted reward functions, the HAIM formulates learning as preference-based policy optimization. Frequent human takeovers in specific contexts (e.g., intersections, merges) are treated as implicit indicators of policy failure, shaping cost signals or trajectory ranking. Meanwhile, physics-based models act as behavioral constraints to ensure learned policies remain safe, interpretable, and socially compliant. This design improves sample efficiency, reduces unsafe exploration, and fosters human trust in the AI system.

\subsubsection{AI as Human Mentor}
In parallel, the AIHM enables AI to function as a real-time coach for human drivers. It leverages Physics-Enhanced Residual Learning (PERL) \cite{long2025physics,sheng2024traffic} to generate optimal driving paths that consider vehicle dynamics, safety margins, and individual driving styles \cite{sheng2024ego}. These reference trajectories are visualized in real time via VR or in-vehicle displays and are continuously updated based on driver performance. AIHM evaluates drivers using metrics such as path deviation, response latency, control stability, and situational awareness. Personalized feedback is delivered through annotated replays, visual heatmaps, and AI-generated verbal summaries.

A key innovation of AIHM is the use of generative AI for scenario customization \cite{sheng2025talk2traffic}. Based on performance analytics, the system dynamically generates targeted training tasks—such as emergency stops or lane changes—to address specific weaknesses. The level of guidance is continuously adjusted using real-time physiological and behavioral signals: when elevated stress levels (e.g., increased heart rate, frequent steering corrections) are detected, the system reduces scenario complexity and provides calming feedback. Conversely, as the driver demonstrates improved performance, the system introduces more challenging conditions to encourage continued skill development \cite{sheng2025curricuvlm}.

\subsection{Digital Twin Framework}
AI algorithms trained in simulation often fail to generalize to real-world traffic due to the lack of environmental fidelity. To address this sim-to-real gap and ensure practical applicability, Sky-Drive introduces a DT framework that creates dynamic, high-fidelity replicas of real transportation systems. 

As illustrated in Fig.\ref{fig2} (a), the DT framework consists of two core components: data integration and virtual environment construction. The multi-source data integration layer fuses static and real-time inputs from traffic cameras, loop detectors, connected vehicle telemetry, GPS traces, historical traffic records, and high-definition maps collected using lab-developed AVs equipped with LiDAR and radar. These inputs undergo temporal alignment, spatial correlation, and feature extraction to ensure semantic consistency across sources. 

The virtual environment is built on CARLA and Unreal Engine and integrates real-time sensor data and computer vision models to detect and track road users for both rendering and trajectory prediction. By employing video recognition and object tracking models, the system reconstructs road user trajectories and maps them into the digital replica, enabling visual analytics, risk prediction, and event replay. Sky-Drive has implemented a pilot deployment of this framework along the Flex Lane on the Beltline in Dane County, Wisconsin. The DT ingests real-time feeds from WisDOT 511 and historical records from WisTransPortal, enabling dynamic reconstruction of traffic states and generation of predictive insights.

\section{Sky-Drive Use Case}
\label{Application}

\begin{figure}[!t]
\centering
\includegraphics[width=0.49\textwidth]{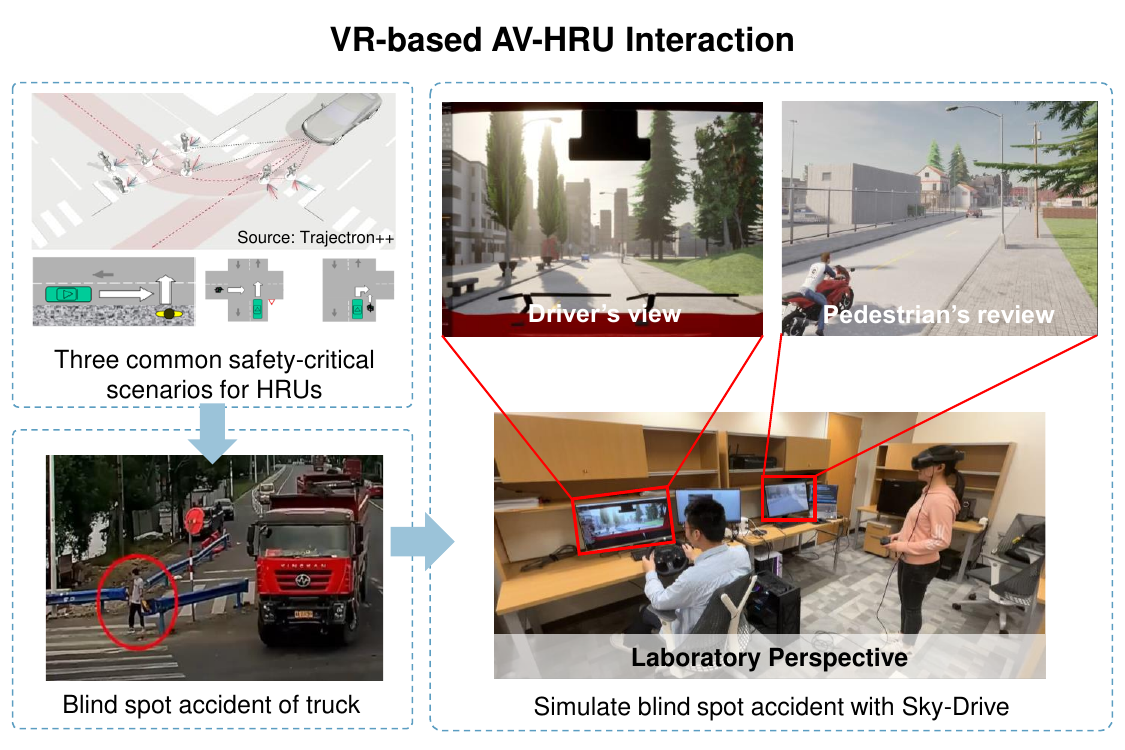}
\caption{VR-based experimental setup for studying AV-HRU interactions at unsignalized intersections.}
\label{fig4}
\end{figure}

\subsection{VR-based AV-HRU Interaction}
Studying interactions between AVs and HRUs is critical for the safe deployment of AV technology in complex urban environments. Although there are reported incidents of AV-HRU conflicts, real-world crash data involving such cases remains scarce. More importantly, collecting such data in real traffic is unsafe, difficult to reproduce, and often restricted by ethical constraints. To address this challenge, Sky-Drive provides a VR-enabled platform for investigating AV-HRU interactions in a controlled, immersive, and data-rich environment. It's a distributed multi-agent simulation architecture that enables synchronized control of multiple agents—across separate terminals and devices—while maintaining real-time coordination. This setup is particularly valuable for modeling high-risk scenarios that are difficult to observe or replicate in the physical world.

As shown in Fig.~\ref{fig4}, we conducted a case study focused on right-turn conflicts at unsignalized intersections—a scenario frequently associated with accidents in urban environments. This study leveraged Sky-Drive's synchronized multi-terminal architecture in a novel experimental setup where human participants experienced the scenario from the pedestrian's perspective through immersive VR, while researchers monitored the decision-making of an AV from a separate terminal and occasionally intervened when necessary. During each interaction, Sky-Drive captured multimodal behavioral data from both the AV and the pedestrian. The VR recorded 3D gaze vectors, eye fixations, and reaction times from the pedestrian, while simultaneously logging control signals, deceleration profiles, and trajectory predictions from the AV. 

This configuration allows researchers to analyze both the physical outcomes (e.g., successful yielding, near-misses, pedestrian hesitation) and the cognitive-emotional states of the human participant, offering insight into how HRUs perceive and respond to AV behavior.

\begin{table*}[!ht]
\centering
\begin{small}
\caption{The performance of PPO, HACO, and HAIM-DRL methods.}
\label{tab2}
\begin{tabular}{@{}cccccc@{}}
\toprule 
Method & \shortstack{Test Safety Violation} {$\downarrow$} & \shortstack{Test Return} {$\uparrow$} & \shortstack{Test Disturbance Rate} {$\downarrow$} & \shortstack{Test Success Rate} {$\uparrow$} & \shortstack{Train Samples} {$\downarrow$} \\
\midrule
PPO &	80.84	& \textbf{1591.00} &	- &	0.35	& 500,000 \\
HACO &	12.14	& 1578.43 &	0.0137 &	0.35	& 8,000 \\
HAIM-DRL (ours) & \textbf{11.25} & 1590.85 & 0.0121 & \textbf{0.38} & \textbf{8,000} \\
\bottomrule
\end{tabular}
\vspace{0.5em}

\parbox{\textwidth}{\raggedright \small \textbf{Note:} The results are based on data reported in \cite{huang2024human}. For detailed definitions of evaluation metrics and descriptions of baseline methods, please refer to the original paper.}
\end{small}
\vspace{-1em}
\end{table*}

\begin{table*}[!t]
\begin{small}
  \caption{Performance comparison with baselines during testing. Mean and standard deviation over 3 seeds. }
  \label{tab4}
  \centering
  \begin{center}
  \renewcommand{\arraystretch}{1.5}
  \begin{tabular}{@{}cccccc@{}}
  \toprule
  Model & Average Speed~$\uparrow$ &  Route Completion~$\uparrow$ & Traveled Distance~$\uparrow$ & Collision Rate~$\downarrow$ & Success Rate~$\uparrow$ \\
  \midrule 
   VLM-SR  &  0.53 {\tiny $\pm$ 0.27} & 0.02 {\tiny $\pm$ 0.00} & 47.9 {\tiny $\pm$ 9.2} & 0.18 {\tiny $\pm$ 0.25} & 0.0 {\tiny $\pm$ 0.0}  \\
   RoboCLIP  &  0.44 {\tiny $\pm$ 0.05} & 0.07 {\tiny $\pm$ 0.03} & 146.3 {\tiny $\pm$ 62.3} & 1.05 {\tiny $\pm$ 0.58} &  0.0 {\tiny $\pm$ 0.0}  \\
   VLM-RM  &  0.20 {\tiny $\pm$ 0.05} & 0.02 {\tiny $\pm$ 0.01} & 35.9 {\tiny $\pm$ 25.8} & \textbf{0.003} {\tiny $\pm$ 0.005} & 0.0 {\tiny $\pm$ 0.0}  \\
   LORD   &  0.17 {\tiny $\pm$ 0.08} & 0.02 {\tiny $\pm$ 0.02} & 45.1 {\tiny $\pm$ 57.1} &  0.02 {\tiny $\pm$ 0.02} & 0.0 {\tiny $\pm$ 0.0} \\
   LORD-Speed  &  18.9 {\tiny $\pm$ 0.36} & 0.87 {\tiny $\pm$ 0.08} & 1783.4 {\tiny $\pm$ 172.8} & 2.80 {\tiny $\pm$ 1.16} & 0.67 {\tiny $\pm$ 0.05}  \\

   VLM-RL (ours)  &  \textbf{19.3} {\tiny $\pm$ 1.29} & \textbf{0.97} {\tiny $\pm$ 0.03} & \textbf{2028.2} {\tiny $\pm$ 96.6} & 0.02 {\tiny $\pm$ 0.03} & \textbf{0.93} {\tiny $\pm$ 0.04}  \\
  \bottomrule
  \end{tabular}
  \end{center}
\end{small}
\vspace{0.5em}

\parbox{\textwidth}{\raggedright \small \textbf{Note:} The best results are marked in \textbf{bold}. The results are based on data reported in \cite{huang2024vlm}. For detailed definitions of evaluation metrics and descriptions of baseline methods, please refer to the original paper.}
\end{table*}

\subsection{HAIM-based Deep Reinforcement Learning}
To validate the HAIM module, we implemented and tested HAIM-DRL \cite{huang2024human}, a reward-free RL approach that enables AI agents to learn driving behavior directly from human interventions. This demonstration serves as a proof-of-concept for the HAIM module's core functionality—leveraging real-time human feedback to guide policy learning—within the multi-agent, simulation-rich environment of Sky-Drive.

Sky-Drive enables HAIM-DRL by detecting and recording steering takeovers, synchronized with vehicle state and surrounding scene context. Within its multi-agent traffic simulation environment, human participants intervene when dissatisfied with the AV’s behavior (e.g., aggressive merging, unsafe following), implicitly indicating suboptimal actions. These interventions are used to construct preference comparisons between pre- and post-takeover trajectories, allowing the agent to identify and avoid human-disapproved actions and iteratively refine its driving policy \cite{huang2024human}. 

Mathematically, the HAIM-DRL can be defined as follows: 
\begin{equation}
\pi^{*}_{\text{AV}} = \arg\min_{\pi_{\text{AV}}} \mathbb{E}_{s_t \sim d_{\pi_{\text{AV}}}} \left[ \mathcal{L}\left( \pi_{\text{AV}}(\cdot \mid s_t), \pi_{\text{human}}(\cdot \mid s_t) \right) \right],
\end{equation}
where \( d_{\pi_{\text{AV}}} \) represents the state distribution induced by the agent's policy \( \pi_{\text{AV}} \), and \( \mathcal{L}(\cdot, \cdot) \) is a measure of discrepancy (e.g., KL divergence). By minimizing this discrepancy over the state distribution, the AI agent is encouraged to align its behavior with human preferences.

The actual trajectory during the training process is determined by the mixed behavior policy:
\begin{equation}
\pi_{\text{mix}}(a \mid s) = \pi_{\text{AV}}(a \mid s)(1 - I(s, a)) + \pi_{\text{human}}(a \mid s)F(s)
\end{equation}
where $F(s) = \int_{a' \notin A_{\eta}(s)} \pi_{\text{AV}}(a' \mid s) \, da'$ represents the probability of the agent selecting an action that would be rejected by the human. \( I(s, a) \) is an indicator function that equals 1 if the human rejects the agent action and 0 otherwise.

The overall learning objective of HAIM-DRL is specifically designed as \cite{huang2024human}:
\begin{equation}
\begin{aligned}
\max_{\pi} \; \mathbb{E} \big[ \; 
& \psi \hat{Q}(s_t, a_t^{\text{AV}}) 
- \alpha \log \pi_{\text{AV}}(a_t^{\text{AV}} \mid s_t; \theta) \\
& - \beta Q^{\text{EX}}(s_t, a_t^{\text{AV}}) 
- \varphi Q^{\text{IM}}(s_t, a_t^{\text{AV}}) \; \big].
\end{aligned}
\label{eq3}
\end{equation}

In the Eq. (\ref{eq3}), the first term guides the agent to align with human-preferred behavior by minimizing the value discrepancy between its own actions and those demonstrated by the human mentor. The second term introduces an entropy regularization factor that encourages the agent to explore diverse strategies. The third term penalizes actions that frequently trigger human takeovers. The fourth term constrains the agent to minimize disturbances to surrounding traffic.

As evidenced by Tab. \ref{tab2}, the HAIM-DRL was successfully implemented and evaluated within the Sky-Drive platform, demonstrating clear advantages over conventional RL methods. Compared with PPO, HAIM-DRL achieves a drastic reduction in safety violations and eliminates the need for large-scale training data, reaching comparable or superior performance with only 8,000 samples. Compared with HACO, which also leverages human interventions, HAIM-DRL further improves test return, reduces disturbance rate, and increases the success rate from 0.35 to 0.38. These results validate Sky-Drive’s capability to support closed-loop human-AI training, enabling efficient, human-aligned policy learning through real-time feedback and preference-driven optimization.

\begin{table*}[!t]
\begin{small}
  \caption{Performance comparison with baselines in the safety-critical test scenarios.}
  \label{tab5}
  \centering
  \begin{center}
  \renewcommand{\arraystretch}{1.5}
  \resizebox{\textwidth}{!}{
  \begin{tabular}{cccccccc}
  \toprule
  \multirow{2}{*}{Model} &  \multirow{2}{*}{\shortstack{Episode\\Reward}} \multirow{2}{*}{$\uparrow$} & \multirow{2}{*}{\shortstack{Road\\Completion (\%)}} \multirow{2}{*}{$\uparrow$} & \multirow{2}{*}{\shortstack{Total\\Distance}} \multirow{2}{*}{$\uparrow$} & \multirow{2}{*}{\shortstack{Crash\\Rate (\%)}} \multirow{2}{*}{$\downarrow$} & \multirow{2}{*}{\shortstack{Average\\Speed}} \multirow{2}{*}{$\uparrow$} & \multirow{2}{*}{\shortstack{Failure-to-\\Success Rate (\%)}} \multirow{2}{*}{$\uparrow$} & \multirow{2}{*}{\shortstack{Success-to-\\Success Rate (\%)}} \multirow{2}{*}{$\uparrow$} \\
  & & & & & & & \\
  \midrule
  
  SAC & 38.4 {\tiny $\pm$ 1.97} & 63.2 {\tiny $\pm$ 1.21} & 40.9 {\tiny $\pm$ 1.34} & 30.5 {\tiny $\pm$ 2.33} & 9.25 {\tiny $\pm$ 0.07}   & 30.4 {\tiny $\pm$ 7.00} & 56.9 {\tiny $\pm$ 15.1}   \\
  PPO & 38.4 {\tiny $\pm$ 0.86} & 62.7 {\tiny $\pm$ 1.05} & 40.0 {\tiny $\pm$ 0.70} & 32.0 {\tiny $\pm$ 2.02} & \textbf{9.94} {\tiny $\pm$ 0.30}  & 26.7 {\tiny $\pm$ 0.89} & 41.7 {\tiny $\pm$ 8.33} \\
  TD3 & 42.4 {\tiny $\pm$ 1.01} & 65.2 {\tiny $\pm$ 1.40} & 42.6 {\tiny $\pm$ 1.26} & 39.7 {\tiny $\pm$ 1.04} & 8.02 {\tiny $\pm$ 0.77}   & 28.6 {\tiny $\pm$ 2.79} & 64.3 {\tiny $\pm$ 21.4} \\
  \midrule 

   CAT  &  42.5 {\tiny $\pm$ 3.95} & 66.6 {\tiny $\pm$ 4.37} & 43.4 {\tiny $\pm$ 3.48} & 32.1 {\tiny $\pm$ 2.08} & 8.36 {\tiny $\pm$ 1.17}  & 35.2 {\tiny $\pm$ 3.44} & 67.5 {\tiny $\pm$ 7.50}  \\
   CLIC  & 39.3 {\tiny $\pm$ 0.72} & 64.3 {\tiny $\pm$ 0.40} & 41.6 {\tiny $\pm$ 0.78} & 26.2 {\tiny $\pm$ 1.17} & 9.21 {\tiny $\pm$ 0.26} & 34.7 {\tiny $\pm$ 2.67} & 61.9 {\tiny $\pm$ 26.9}  \\
  CurricuVLM (ours) & \textbf{48.9} {\tiny $\pm$ 1.53} & \textbf{73.4} {\tiny $\pm$ 1.66} & \textbf{48.4} {\tiny $\pm$ 1.31} & \textbf{25.1} {\tiny $\pm$ 1.17} & 9.45 {\tiny $\pm$ 0.16} & \textbf{39.1} {\tiny $\pm$ 0.66} & \textbf{73.5} {\tiny $\pm$ 21.1}   \\
  \bottomrule
  \end{tabular}
  }
  \end{center}
\end{small}
\vspace{0.5em}

\parbox{\textwidth}{\raggedright \small \textbf{Note:} The best results are marked in \textbf{bold}. The results are based on data reported in \cite{sheng2025curricuvlm}. For detailed definitions of evaluation metrics and descriptions of baseline methods, please refer to the original paper.}
\end{table*}

\subsection{Vision Language Model-Enabled Reinforcement Learning}
To validate Sky-Drive’s capability to support VLM-enabled RL, we implemented the VLM-RL \cite{huang2024vlm}, which integrates pre-trained VLMs with RL to generate semantic reward signals from image observations and natural language goals. This demonstration showcases Sky-Drive’s ability to enable high-level, human-interpretable guidance for safe and efficient autonomous driving.

At the core of VLM-RL is the contrasting language goal (CLG)-as-reward paradigm, which uses pre-trained VLMs to compute semantic similarity between driving states and paired language descriptions. Positive goals (e.g., “the road is clear with no accidents”) and negative goals (e.g., “two cars have collided”) are used to guide the agent’s behavior by comparing how closely its current state aligns with each description. The reward is computed by encoding visual input via CLIP’s image encoder and goals via its text encoder, both mapped into a shared latent space \cite{huang2024vlm}:
\begin{equation}
\begin{aligned}
R_{\text{CLG}}(s) =\; & \alpha \cdot \text{sim}(\text{VLM}_\text{I}(\psi(s)), \text{VLM}_\text{L}(l_{\text{pos}})) \\
                      & - \beta \cdot \text{sim}(\text{VLM}_\text{I}(\psi(s)), \text{VLM}_\text{L}(l_{\text{neg}}))
\end{aligned}
\end{equation}
where \( l_{\text{pos}} \) and \( l_{\text{neg}} \) are the positive and negative language goals, $\text{VLM}_\text{I}$ and $\text{VLM}_\text{L}$ denote the image and language encoders of the pre-trained VLM, \( \psi(s) \) is the visual preprocessing function, and \( \text{sim}(\cdot, \cdot) \) represents cosine similarity. The weights \( \alpha \) and \( \beta \) control the influence of the positive and negative goals, respectively.

To improve reward stability, VLM-RL introduces a hierarchical reward synthesis strategy that combines CLG-based semantic rewards with low-level vehicle state signals such as speed alignment, lane deviation, and directional consistency. The synthesized reward is defined as \cite{huang2024vlm}:
\begin{equation}
R_{\text{synthesis}}(s) = r_{\text{speed}}(s) \cdot f_{\text{center}}(s) \cdot f_{\text{angle}}(s) \cdot f_{\text{stability}}(s)
\end{equation}
where \( r_{\text{speed}}(s) = 1 - \frac{|v - v_{\text{target}}|}{v_{\max}} \) measures speed alignment with respect to the target velocity \( v_{\text{target}} = r_t^{\prime \text{CLG}} \cdot v_{\max} \); \( f_{\text{center}}(s) \) evaluates the vehicle's lateral position relative to the lane center; \( f_{\text{angle}}(s) \) reflects the vehicle's orientation with respect to the road direction; and \( f_{\text{stability}}(s) \) quantifies the temporal consistency of the vehicle’s lateral positioning.

As shown in Tab.~\ref{tab4}, VLM-RL significantly outperforms existing approaches across all key metrics. VLM-RL achieves the highest success rate and route completion, while maintaining a low collision speed of 0.02 km/h—matching the safety level of the most conservative baselines. Unlike existing VLM-based methods, which suffer from overly cautious behavior and near-zero task success, VLM-RL balances safety with efficiency, reaching an average speed of 19.3 km/h and a total driving distance of 2028.2 m. Compared to LLM-based methods such as Revolve, VLM-RL maintains comparable success and completion rates while drastically reducing collision speed. The successful implementation of VLM-RL within the Sky-Drive platform validates its capability to support large-scale, multimodal policy learning grounded in human-understandable semantics.

\subsection{Personalized Safety-Critical Curriculum Learning }

To validate Sky-Drive’s capability to support adaptive scenario generation and curriculum learning, we implement the CurricuVLM \cite{sheng2025curricuvlm}. CurricuVLM integrates VLMs to enable personalized, safety-critical training scenarios tailored to the evolving weaknesses of autonomous driving agents.

The core innovation of CurricuVLM lies in bridging the gap between scenario generation and policy learning. By continuously monitoring agent performance, the framework identifies failure patterns through a two-stage behavior analysis pipeline: VLMs are first used to extract rich visual descriptions of unsafe events, which are then interpreted by a GPT-4o-based analyzer to uncover behavioral limitations. This process enables semantic understanding of critical driving mistakes without manual annotation.

Based on the analysis, scenario generation is formulated as a conditional trajectory generation problem:
\begin{equation}
P(Y^{AV}, Y^{BV}|I, X)
\end{equation}
where $X$ encodes historical context (e.g., maps, past trajectories), $I$ contains semantic insights from behavior analysis, and $Y^{AV}$, $Y^{BV}$ denote future trajectories of the ego and background vehicles, respectively. 
  
The framework optimizes $Y^{BV}$ to generate targeted, informative interactions via:
\begin{equation}
\begin{aligned}
Y^{\text{BV}*} = & \arg\max_{Y^{\text{BV}}} \; P(Y^{\text{BV}} \mid X) \cdot \\
& \sum_{Y^{\text{AV}} \sim \mathcal{Y}(\pi)} P(Y^{\text{AV}} \mid Y^{\text{BV}}, X) \cdot P(I \mid Y^{\text{AV}}, Y^{\text{BV}})
\end{aligned}
\end{equation}
where \( \mathcal{Y}(\pi) \) denotes the trajectory distribution induced by the current policy \( \pi \), and \( P(I \mid Y^{\text{AV}}, Y^{\text{BV}}) \) measures how well the generated scenario aligns with the identified behavioral insight. This formulation encourages the background vehicle behavior to induce targeted policy responses from the AV agent, forming the foundation for automated curriculum construction.

\begin{figure}
  \centerline{\includegraphics[width=0.49\textwidth]{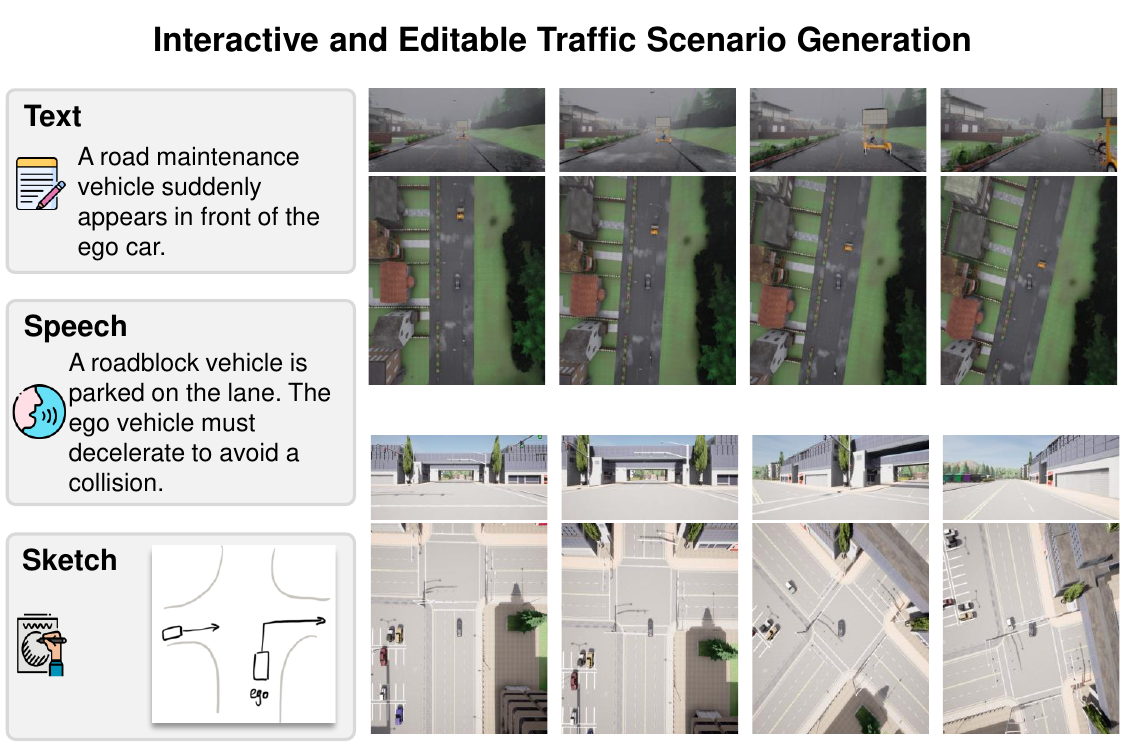}}
  \caption{Qualitative examples. Each scenario is downsampled to four frames for visualisation.}
  \label{fig5}
\end{figure}

As shown in Tab.~\ref{tab5}, CurricuVLM achieves the best overall performance across all key metrics, demonstrating both high safety and training effectiveness. In terms of task performance, CurricuVLM achieves the highest episode reward (48.9) and road completion rate (73.4\%), while maintaining a low crash rate (25.1\%), outperforming baselines such as CAT and CLIC. It also records the highest total driving distance (48.4m) and failure-to-success rate (39.1\%), indicating superior adaptability to previously failed scenarios. Meanwhile, its success-to-success rate (73.5\%) reflects strong behavioral consistency and learning stability. These results validate that CurricuVLM not only enhances policy robustness under long-tail safety-critical scenarios, but also integrates seamlessly into Sky-Drive’s human-AI collaboration mechanism. Some qualitative examples of generated scenarios are illustrated in Fig.~\ref{fig5}. 

\subsection{Accident Data Replay}
To validate Sky-Drive’s capability in supporting real-world accident reconstruction and analysis, we implemented an accident data replay framework that enables systematic reproduction of traffic collisions within Sky-Drive environment. This framework addresses a fundamental need in autonomous driving development: understanding and learning from real accidents in a safe, repeatable, and controlled setting.

\begin{figure}
  \centerline{\includegraphics[width=0.49\textwidth]{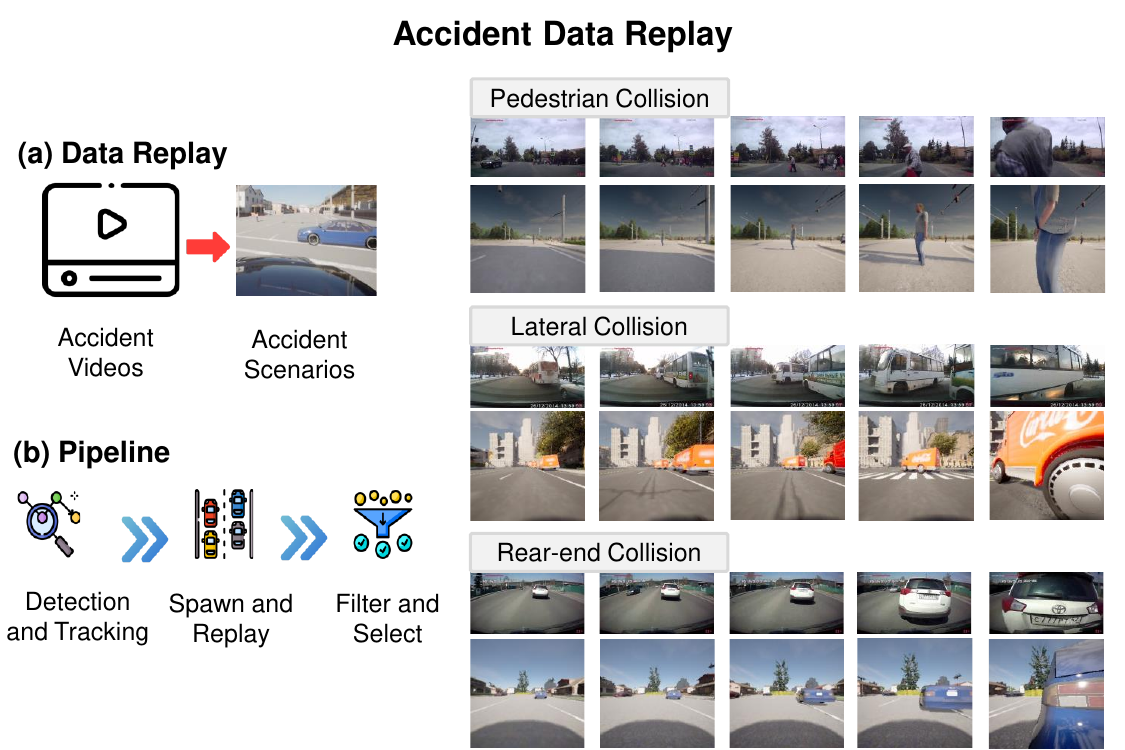}}
  \caption{Accident data replay framework for systematic traffic incident analysis.}
  \label{fig6}
\end{figure}

The replay pipeline centers around CenterTrack \cite{zhou2020tracking}, an advanced multi-object tracking algorithm used to extract object trajectories from raw accident video footage. These 2D trajectories are then mapped into 3D space and replayed in Unreal Engine through Sky-Drive’s integrated simulation backend. As shown in Fig. \ref{fig6}, the reconstructed scenes preserve key dynamics such as vehicle positions, speeds, and interaction sequences, along with contextual factors like road layout and weather. Specially, Sky-Drive incorporates a robust reconstruction validation process to ensure fidelity. A procedural matching algorithm selects the most appropriate simulation maps based on road topology and scene semantics. A built-in quality assessment module scores the visual and kinematic similarity between the replayed and original sequences, flagging low-fidelity cases for refinement. The framework also supports unsupervised domain adaptation to improve trajectory accuracy, while offering manual editing tools when needed to ensure precise alignment with real-world footage.

This replay capability enables several downstream applications: (i) It provides a safe testbed for analyzing accident causation, enabling the development of improved safety mechanisms and behavior prediction models; (ii) It allows RL agents to be trained and evaluated on real-world edge cases, significantly enhancing their robustness in critical scenarios; (iii) It supports regulatory compliance and post-incident investigation by producing detailed, verifiable accident reconstructions. Through this integration, Sky-Drive enables scalable, high-fidelity replay of accident scenarios, positioning itself as a comprehensive platform for evaluating autonomous systems under rare, safety-critical conditions that are otherwise difficult or unsafe to replicate.

\section{Future Enhancements}
\label{future}

\subsection{Foundation Models Integration}

\subsubsection{Multimodal Behavioral Understanding}
Interpreting human behavioral signals in a unified, context-aware manner remains an open challenge. Future iterations of Sky-Drive will leverage LLMs and VLMs to perform cross-modal reasoning across physiological, visual, verbal, and control-based modalities. For example, an elevated heart rate, downward gaze, and a quick verbal cue like “too fast” may collectively indicate the driver’s discomfort with vehicle acceleration. A more nuanced phrase such as “I feel a bit uneasy because the car accelerates too quickly” can be semantically aligned with facial tension and biometric signals like heart rate variability. By combining these signals in the context of traffic density, road geometry, and interaction with nearby vehicles, Sky-Drive can construct rich behavioral profiles far beyond what single-modality systems can achieve. This capability will support personalized feedback generation, trust modeling, and adaptive control within the HAIM and AIHM modules.

\subsubsection{Personalized Autonomous Driving}
\begin{figure}
  \centerline{\includegraphics[width=0.49\textwidth]{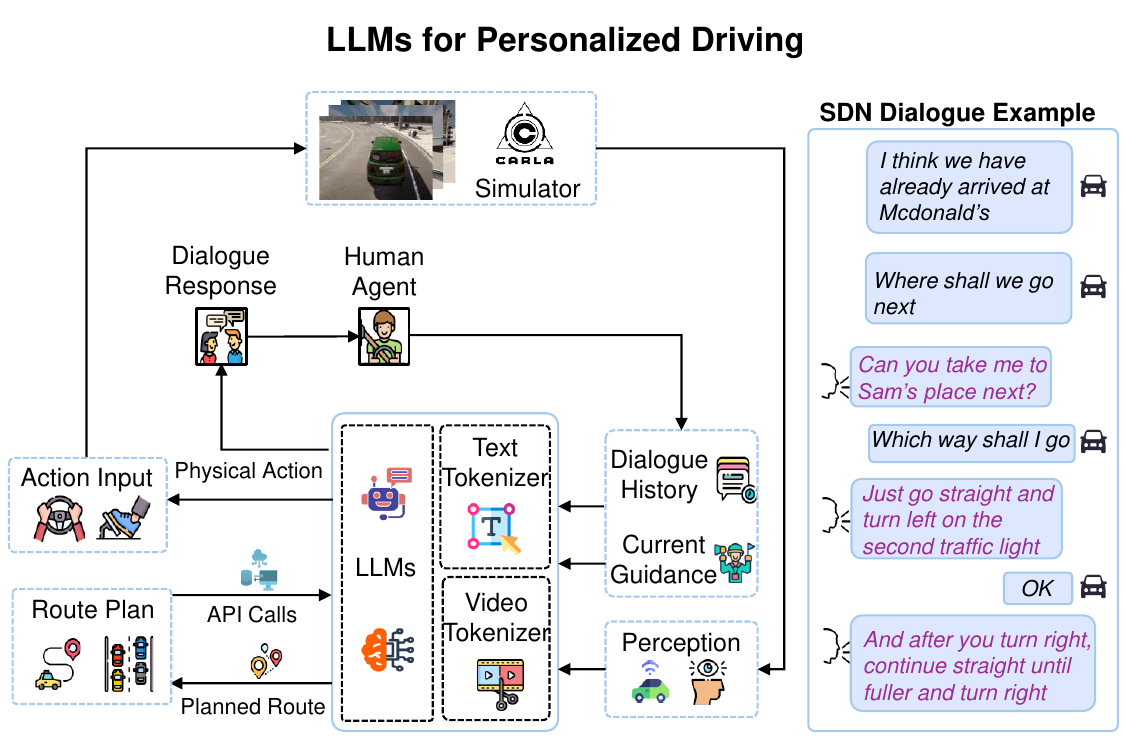}}
  \caption{LLM-based system enabling personalized autonomous driving.}
  \label{fig7}
\end{figure}

As shown in Fig \ref{fig7}, Sky-Drive will implement a LLM-based system that enables personalized autonomous driving through natural language interaction \cite{xu2025personalizing}. Specifically, the system will integrate three core modules: a visual encoder to process real-time camera feeds, an LLM to interpret language inputs, and a route planning module to generate executable commands based on Sky-Drive’s maps. To ensure robustness, Sky-Drive will use a three-stage training pipeline. The first stage uses the BDD-X dataset \cite{xu2017end} to align visual and linguistic representations. The second stage fine-tunes language understanding via LoRA techniques on the SDN dataset \cite{ma2022dorothie}. The final stage incorporates data generated within the Sky-Drive simulation environment to adapt model responses to realistic driving tasks. This integration allows drivers to provide real-time feedback such as “slow down a bit here” or “take the next left,” and have the vehicle respond accordingly. In the long term, this capability will support personalized, explainable, and user-aligned driving experiences. 

\subsubsection{ Traffic Brains }
Sky-Drive will position foundation models as intelligent “traffic brains” that govern decision-making in complex, multi-agent traffic environments. While general-purpose models such as Qwen \cite{Qwen2VL}, GPT \cite{openai2023gpt4}, and Llama \cite{dubey2024llama} exhibit strong language and reasoning abilities, they will require domain-specific adaptation to meet the demands of autonomous driving. To address this challenge, Sky-Drive will leverage transportation-specific datasets—including LMDrive \cite{shao2024lmdrive}, CCD \cite{BaoMM2020}, DoTA \cite{yao2022dota}, and DriveCoT \cite{wang2024drivecot}—to fine-tune pre-trained foundation models. This fine-tuning pipeline is designed to enhance the model’s ability to handle dynamic scenario adaptation, hierarchical reasoning, and multitask decision-making, including generating safe control actions (e.g., steering, throttle, and braking) and predicting critical safety metrics such as time-to-collision (TTC). The refined models will be deployed within the Sky-Drive simulation environment to enable coordinated behavior across AVs and other components, facilitating holistic control and system-level optimization. 

\subsection{Hardware-in-the-Loop}

\subsubsection{Simulation-to-Reality Integration }
As shown in Fig. \ref{fig1} (e), the center of the HIL framework is a Ford E-Transit electric van retrofitted with fully automated driving capabilities. The vehicle is equipped with a comprehensive sensor suite—including LiDAR, radar, high-resolution cameras, and OxTS navigation units—and operates on a drive-by-wire system connected to an industrial-grade computing rack. Sky-Drive connects its real-time simulation environment with the physical vehicle through standardized ROS interfaces to realize hardware-in-the-loop testing. Simulated sensor streams (e.g., camera and LiDAR) are fed into the vehicle’s onboard systems, which process the data and return control commands (e.g., throttle, brake, steering) to the simulator. This closed-loop integration allows developers to evaluate real hardware behavior under diverse and repeatable traffic scenarios, including safety-critical scenarios that are difficult or unsafe to test on public roads. By testing algorithms in simulation before deployment, Sky-Drive can bridge the sim-to-real gap and ensure that experimental validation can be conducted safely without exposing operators or the public to real-world risks.




\subsubsection{Remote Driving and Data Collection} 
The HIL framework also establishes a solid foundation for developing and testing teleoperated driving. Teleoperated driving allows humans (teleoperators) to remotely control vehicles, particularly in challenging scenarios, complementing fully/highly autonomous solutions. It is one of the important use cases of V2X communication, specified in the 3GPP standards~\cite{3gpp_tr_22886,huang2024toward,you2024v2x}. Sky-Drive's ROS integration enables wireless connectivity between the testbed vehicle and the human-in-the-loop framework via cellular or satellite networks, such as 5G. To support this, Sky-Drive integrates a 5G mmWave mobile hotspot (i.e., NETGEAR Nighthawk M6 Pro) to enable remote control and real-time data collection. This setup allows researchers to control vehicles without being physically onboard, ensuring operator safety during high-risk scenarios. Meanwhile, it supports remote collection of human behavioral data (e.g., steering patterns and pedal inputs), even when the vehicle and experiment coordinators are located in different cities or campuses.



\section{Conclusions}
\label{Conclusions}

This paper presented \textbf{Sky-Drive}, a distributed multi-agent simulation platform designed for socially-aware autonomous driving and human-AI collaboration in future transportation systems. Unlike existing simulators that primarily focus on validating single-vehicle performance, Sky-Drive addresses the emerging need to explore complex interactions in mixed traffic environments where various intelligent agents must align with human preferences and societal norms.

Sky-Drive introduces several key innovations: (a) a distributed multi-agent architecture enabling synchronized simulation across multiple terminals, allowing independent agent control while maintaining shared environmental states; (b) a multi-modal human-in-the-loop framework integrating diverse sensors to capture comprehensive behavioral data; (c) a novel human-AI collaboration mechanism to facilitate bidirectional knowledge exchange; and (d) a digital twin framework creating high-fidelity virtual replicas of real transportation systems. The platform's effectiveness has been demonstrated through multiple application cases, including VR-based vulnerable road user interactions, physics-enhanced reinforcement learning with human feedback, vision-language model-enabled reinforcement learning, personalized curriculum learning, and accident data replay. These applications show Sky-Drive's potential to advance autonomous driving research beyond traditional metrics of safety and efficiency toward more socially aware and human-aligned behavior.

To further enhance Sky-Drive's capabilities, we have outlined two major planned functionalities: (i) the integration of foundation models to support multimodal behavior understanding, personalized driving, and system-level optimization via traffic brains; and (ii) a hardware-in-the-loop framework via ROS integration to enable direct validation of algorithms on physical vehicles. These future enhancements will bridge the gap between simulation and reality, ensuring that algorithms are safely evaluated in real-world environments. As autonomous driving technology continues to evolve, Sky-Drive provides a robust platform for ensuring that future transportation systems are not only safe and efficient but also socially aware and aligned with human expectations.

\section*{Acknowledgments}
This work was funded by the U.S. Department of Transportation, Award \#69A3552348305.  In addition, this research was supported by grants from NVIDIA and utilized NVIDIA RTX PRO 6000 Blackwell Max-Q Workstation Edition GPUs and A100 GPU-Hours on Saturn Cloud.

\bibliographystyle{IEEEtran} 
\bibliography{IEEEabrv,ref_abb} 

\vfill

\end{document}